\pdfoutput=1

\documentclass[11pt]{article}

\usepackage[]{EMNLP2022}

\usepackage{times}
\usepackage{latexsym}

\usepackage[T1]{fontenc}

\usepackage[utf8]{inputenc}
\usepackage{CJKutf8}

\usepackage{microtype}

\usepackage{inconsolata}
\usepackage{graphicx}
\usepackage{booktabs}
\usepackage{color}
\usepackage{multirow}
\usepackage{multicol}
\usepackage{subfigure}
\usepackage{makecell}
\usepackage{enumerate}
\newcommand{\DatasetName}{NaCGEC}
\newcommand{\MethodName}{CLG}

\usepackage{xcolor}
\usepackage{epigraph}
\title{Linguistic Rules-Based Corpus Generation for Native Chinese Grammatical Error Correction}

\author{
Shirong Ma$^{1}$\thanks{ $^*$ indicates equal contribution. Work is done during Yinghui's internship at Tencent Cloud Xiaowei.},~Yinghui Li$^{1*}$,~Rongyi Sun$^{1}$,~Qingyu Zhou$^{2}$,~Shuling Huang$^{1}$,~Ding Zhang$^{1}$,\\~\textbf{Yangning Li}$^{1}$,~\textbf{Ruiyang Liu}$^{4}$,~\textbf{Zhongli Li}$^{2}$, ~\textbf{Yunbo Cao}$^{2}$, \textbf{Hai-Tao Zheng}$^{1,3}$\thanks{ $^{\dagger}$ Corresponding author: Hai-Tao Zheng and Ying Shen. (E-mail: zheng.haitao@sz.tsinghua.edu.cn, sheny76@mail.sysu.edu.cn)}, \textbf{Ying Shen}$^{5\dagger}$\\
        $^{1}$Tsinghua Shenzhen International Graduate School, Tsinghua University \\ 
        $^{2}$Tencent Cloud Xiaowei, $^{3}$Peng Cheng Laboratory \\
        $^{4}$Department of Computer Science and Technology, Tsinghua University\\
        $^{5}$School of Intelligent Systems Engineering, Sun-Yat Sen University\\
        \texttt{\{masr21,liyinghu20\}@mails.tsinghua.edu.cn}
}

\begin{document}
\maketitle
\begin{abstract}
Chinese Grammatical Error Correction (CGEC) is both a challenging NLP task and a common application in human daily life.
Recently, many data-driven approaches are proposed for the development of CGEC research.
However, there are two major limitations in the CGEC field:
First, the lack of high-quality annotated training corpora prevents the performance of existing CGEC models from being significantly improved.
Second, the grammatical errors in widely used test sets are not made by native Chinese speakers, resulting in a significant gap between the CGEC models and the real application.
In this paper, we propose a linguistic rules-based approach to construct large-scale CGEC training corpora with automatically generated grammatical errors.
Additionally, we present a challenging CGEC benchmark derived entirely from errors made by native Chinese speakers in real-world scenarios.
Extensive experiments\footnote{Our dataset and source codes are available at \url{https://github.com/masr2000/CLG-CGEC}.} and detailed analyses not only demonstrate that the training data constructed by our method effectively improves the performance of CGEC models, but also reflect that our benchmark is an excellent resource for further development of the CGEC field.
\end{abstract}

\section{Introduction}
\epigraph{In the field of theoretical linguistics, the native speaker is the authority of the grammar.}{\textit{- Noam Chomsky}}

\noindent Chinese Grammatical Error Correction (CGEC) aims to automatically correct grammatical errors that violate language rules and converts the noisy input texts to clean output texts~\citep{DBLP:journals/corr/abs-2005-06600}. 
In recent years, CGEC has attracted more and more attention from NLP researchers due to its broader applications in all kinds of daily scenarios and downstream tasks~\citep{DBLP:conf/www/DuanH11, DBLP:conf/ltconf/KubisVWZ19, omelianchuk-etal-2020-gector}. 

With the progress of deep learning, data-driven methods based on neural networks, e.g., Transformer~\cite{DBLP:conf/nips/VaswaniSPUJGKP17}, have become the mainstream for CGEC~\citep{DBLP:conf/aaai/ZhaoW20, tang2021chinese, DBLP:journals/corr/abs-2204-10994, DBLP:journals/corr/abs-2205-10884}.
However, we argue that there are still two problems in CGEC: 
(1) \textbf{For model training}, owing to the limited number of real sentences containing grammatical errors, the long-term lack of high-quality annotated training corpora hinders many data-driven models from exercising their capabilities on the CGEC task.
(2) \textbf{For model evaluation}, the widely used benchmarks such as NLPCC~\cite{DBLP:conf/nlpcc/ZhaoJS018} and CGED~\cite{rao-etal-2018-overview, rao-etal-2020-overview} are all derived from the grammatical errors made by foreign Chinese learners (i.e., L2 learners) in their process of learning Chinese, the gap between the language usage habits of L2 learners and Chinese native speakers makes the performance of the CGEC models in real scenarios unpredictable. 

As illustrated in Table~\ref{tab:example}, the samples of NLPCC and CGED are both from L2 learners, so their sentence structures are relatively short and simple. More crucially, the grammatical errors in these samples are very obvious and naive. On the other hand, in the third example, the erroneous sentence is fluent on the whole, which shows that the grammatical errors made by native speakers are more subtle, that is, they are actually in line with the habit of speaking in people's daily communication, but they do not conform to linguistic norms. \emph{Therefore, in the broader scenarios of Chinese usage besides foreigners learning Chinese, we believe that wrong sentences made by native speakers are more valuable and can better evaluate the model performance than errors made by L2 learners.}

\begin{CJK*}{UTF8}{gbsn}
\begin{table*}[t]
\centering
\small

\begin{tabular}{l|l}
\toprule
\multirow{5}{*}{NLPCC} & \textbf{Incorrect:} 那个消息给我\textcolor{red}{打}冲击。\\
 & \textbf{Translation:} The news gave me a \textcolor{red}{hit} shock.\\
 & \textbf{Correct:} 那个消息给我\textcolor{red}{很大}冲击。 \\ 
 & \textbf{Translation:} The news gave me a \textcolor{red}{big} shock. \\
 & \textbf{Source: L2 learners} \\
 \midrule
\multirow{5}{*}{CGED} & \textbf{Incorrect:} 他\textcolor{red}{非常}被日本的风景吸引了。 \\
 & \textbf{Translation:} He was \textcolor{red}{very} attracted by the Japanese landscape. \\
 & \textbf{Correct:} 他\textcolor{red}{深深地}被日本的风景吸引了。 \\ 
 & \textbf{Translation:} He was \textcolor{red}{deeply} attracted by the Japanese landscape. \\
 & \textbf{Source: L2 learners} \\
 \midrule
\multirow{5}{*}{\makecell[c]{Native\\Error}} & \textbf{Incorrect:} 站在\textcolor{red}{即将到来的}2017年的起跑线上，我们\textcolor{red}{不由得}情不自禁地感到自豪和喜悦。 \\
 & \textbf{Translation:} Standing at the starting line of \textcolor{red}{upcoming} 2017, we cannot \textcolor{red}{have to} help feeling proud and joyful.\\
 & \textbf{Correct:} 站在2017年的起跑线上，我们情不自禁地感到自豪和喜悦。\\
 & \textbf{Translation:} Standing at the starting line of 2017, we cannot help feeling proud and joyful. \\
 & \textbf{Source: Native speakers} \\
\bottomrule
\end{tabular}%
\caption{Examples of grammatical errors from NLPCC, CGED and Chinese native speakers respectively.}
\label{tab:example}
\end{table*}

\textbf{To alleviate the dilemma of missing large-scale training data}, we propose \MethodName{}, a novel approach based on Chinese linguistic rules that automatically constructs high-quality ungrammatical sentences from grammatical corpus. 
Specifically, according to authoritative linguistic books~\cite{huangborong2011, shaojingmin2016}, we divide Chinese grammatical errors into 6 categories, and design detailed grammatical rules to generate corresponding erroneous sentences according to the characteristics of their respective errors. Our divided 6 error types are: 
\emph{Structural Confusion}, 
\emph{Improper Logicality}, 
\emph{Missing Component}, 
\emph{Redundant Component}, 
\emph{Improper Collocation}, 
and \emph{Improper Word Order}. 
Different from traditional data augmentation, ungrammatical sentences generated by \MethodName{} are more closely matching actual errors that Chinese native speakers would make.
Benefiting from our proposed \MethodName{}, high-quality and large-scale training samples are automatically constructed with annotated error types. 
Moreover, \textbf{to fill the gap between existing benchmarks and practical applications}, we collect a test dataset containing grammatical errors made by native Chinese speakers in real scenarios, named \DatasetName{}, which will be a more challenging benchmark and a meaningful resource to facilitate further development of CGEC.

We conduct extensive experiments to demonstrate the effectiveness of \MethodName{} and the challenge of \DatasetName{}.
Quantitative experiments show that the model trained on our generated corpus performs better than that trained on traditional CGEC datasets. And compared with general data augmentation methods, the training data obtained by \MethodName{} brings larger performance improvements.
In addition, qualitative analyses illustrate that it is more difficult for well-educated Chinese native speakers to identify grammatical errors in \DatasetName{} than in previous existing benchmarks, which indicates that errors in \DatasetName{} are closer to the real mistakes that native speakers would make in their daily life. 
We believe that our proposed corpus generation approach and benchmark can greatly contribute to the development of CGEC methods.

\section{Related Work}
\subsection{CGEC Resources}
Compared with English Grammatical Error Correction (EGEC), data resources for CGEC are still lacking. 
The NLPCC~\cite{DBLP:conf/nlpcc/ZhaoJS018} provides a test set containing 2K sentences and a large-scale dataset collected from the Lang-8 website for training model.
The CGED~\cite{rao-etal-2018-overview, rao-etal-2020-overview} is an evaluation dataset focusing on error diagnosis which contains 5K sentences from HSK corpus~\cite{cui2011principles, zhang2013design}.
The entire HSK corpus can be also utilized for model training. 
The YACLC~\cite{DBLP:journals/corr/abs-2112-15043} collects and annotates 32K sentences from Lang-8 to construct a CGEC dataset.
The latest MuCGEC~\cite{DBLP:journals/corr/abs-2204-10994} selects and re-annotates sentences from the NLPCC, CGED, and Lang-8 corpora to obtain a multi-reference evaluation dataset with 7K sentences.
To the best of our knowledge, no existing resources focus on the grammatical errors made by Chinese native speakers, all of these above-mentioned datasets originate from errors made by L2 learners.

\subsection{CGEC Methods}
CGEC can be considered as a seq2seq task. 
Some existing works employ CNN-based~\cite{DBLP:conf/nlpcc/RenYX18} or RNN-based~\cite{DBLP:conf/nlpcc/ZhouLLBXL18} models to resolve the CGEC task.
Most later work~\cite{wang2020chinese, tang2021chinese, DBLP:conf/aaai/ZhaoW20} employs the Transformer~\cite{DBLP:conf/nips/VaswaniSPUJGKP17} model which has been a great success in Machine Translation. Those studies also propose some data augmentation approaches to extend the training data for improving the model performance.
Recently, researchers start to treat CGEC as a seq2edit task that iteratively predicts the modification label for each position of the sentence.
Similar to the GECToR~\cite{omelianchuk-etal-2020-gector}, \citet{liang-etal-2020-bert} utilize a seq2edit model for CGEC.
\citet{DBLP:journals/corr/abs-2204-10994} directly adopt GECToR for CGEC and enhances it by using pretrained language models.
TtT~\cite{li-shi-2021-tail} proposes a non-autoregressive CGEC approach by employing the BERT~\cite{devlin-etal-2019-bert} encoder with CRF~\cite{DBLP:conf/icml/LaffertyMP01}.
\citet{DBLP:journals/corr/abs-2205-10884} proposes a sequence-to-action model to resolve the CGEC task, which combines the advantages of both seq2seq and seq2edit approaches.
Unlike previous works, we first focus on the linguistic rules of Chinese grammar and exploit them to automatically obtain high-quality training corpora to improve the performance of CGEC models.

\section{Automatic Corpus Generation and Benchmark Construction}
\subsection{Schema Definition}
According to the authoritative linguistic books~\cite{huangborong2011, shaojingmin2016}, Chinese grammatical errors are categorized into 7 types: 
\emph{Structural Confusion}, 
\emph{Improper Logicality}, 
\emph{Missing Component}, 
\emph{Redundant Component}, 
\emph{Improper Collocation}, 
\emph{Improper Word Order} 
and \emph{Ambiguity}. 
It is worth noting that the errors of ambiguity are often caused by the lack of context information. So if we want the model to correct such errors, we must have enough additional knowledge besides grammar, which is beyond the essence of the CGEC task. Therefore, we do not consider this type of error. 
In addition, there is a class of common errors, i.e., spelling errors which are mainly caused by various confusing characters with similar strokes/pronunciations~\cite{li-etal-2022-past}.
But \citeauthor{wang-etal-2018-hybrid} have comprehensively studied how to automatically generate large-scale training data containing spelling errors. How to automatically generate high-quality training data is one of our core contributions, so we also don't need to focus on spelling errors in our study. 

From the linguistic point of view, the schema of these 6 error types is explained as follows:
\begin{enumerate}[(1)]
\item \textbf{Structural Confusion} (结构混乱) means to mix two or more different syntactic structures in one sentence, which results in confusing sentence structure.

\item \textbf{Improper Logicality} (不合逻辑) represents that the meaning of a sentence is inconsistent or does not conform to objective reasoning.

\item \textbf{Missing Component} (成分残缺) means that the sentence structure is incomplete and some grammatical components are missing.

\item \textbf{Redundant Component} (成分冗余) refers to an addition of unnecessary words or phrases to a well-structured sentence.

\item \textbf{Improper Collocation} (搭配不当) is that the collocation between some components of a sentence does not conform to the structural rules or grammatical conventions of Chinese.

\item \textbf{Improper Word Order} (语序不当) mainly refers to the ungrammatical order of words or clauses in a sentence.
\end{enumerate}
Some example sentences containing various grammatical errors and the corresponding corrections are presented in Table~\ref{tab:example2}. These examples are all selected from our proposed benchmark which will be mentioned in Section~\ref{sec:benchmark}.

Next, we will further introduce the process of our proposed \MethodName{} and the details of \DatasetName{}.

\begin{table*}[t]
\small
\centering
\begin{tabular}{l|l}
\toprule
\multirow{4}{*}{\makecell[c]{Structural\\Confusion}} & \textbf{Incorrect:} 食用水果前应该洗净削皮\textcolor{red}{较为安全}。 \\
 & \textbf{Translation:} Fruit should be washed and peeled before eating \textcolor{red}{is safer}. \\
 & \textbf{Correct:} 食用水果前应该洗净削皮。 \\
 & \textbf{Translation:} Fruit should be washed and peeled before eating. \\
\midrule

\multirow{4}{*}{\makecell[c]{Improper\\Logicality}} & \textbf{Incorrect:} 集团向社会各界人士、\textcolor{red}{沿途村庄百姓}表示歉意。 \\
 & \textbf{Translation:} The group apologizes to people from all walks of life \textcolor{red}{and villagers along the way}. \\
 & \textbf{Correct:} 集团向社会各界人士表示歉意。 \\
 & \textbf{Translation:} The group apologizes to people from all walks of life. \\
\midrule

\multirow{4}{*}{\makecell[c]{Missing\\Component}} & \textbf{Incorrect:} 该节目成功地实现了收视冠军。 \\
 & \textbf{Translation:} The program has successfully achieved a ratings champion. \\
 & \textbf{Correct:} 该节目成功地实现了收视冠军\textcolor{red}{的目标}。 \\
 & \textbf{Translation:} The program has successfully achieved \textcolor{red}{the goal of being} a ratings champion. \\
\midrule

\multirow{4}{*}{\makecell[c]{Redundant\\Component}} & \textbf{Incorrect:} 昨天是转会\textcolor{red}{截止日期}的最后一天。 \\
 & \textbf{Translation:} Yesterday was the last day of the transfer \textcolor{red}{deadline}. \\
 & \textbf{Correct:} 昨天是转会的最后一天。\\
 & \textbf{Translation:} Yesterday was the last day of the transfer. \\
\midrule

\multirow{4}{*}{\makecell[c]{Improper\\Collocation}} & \textbf{Incorrect:} 丝绸之路\textcolor{red}{开拓}了千古传诵的壮美篇章。\\
 & \textbf{Translation:} The Silk Road has \textcolor{red}{opened} a magnificent chapter that has been passed down through the ages. \\
 & \textbf{Correct:} 丝绸之路\textcolor{red}{谱写}了千古传诵的壮美篇章。\\
 & \textbf{Translation:} The Silk Road has \textcolor{red}{written} a magnificent chapter that has been passed down through the ages. \\
\midrule

\multirow{4}{*}{\makecell[c]{Improper\\Word Order}} & \textbf{Incorrect:} 学校\textcolor{red}{三个月内要求每名学生}完成20个小时的义工服务。\\
 & \textbf{Translation:} The school \textcolor{red}{in three months} \textcolor{red}{requires each student} to complete 20 hours of volunteer service. \\
 & \textbf{Correct:} 学校\textcolor{red}{要求每名学生三个月内}完成20个小时的义工服务。 \\
 & \textbf{Translation:} The school \textcolor{red}{requires each student} to complete 20 hours of volunteer service \textcolor{red}{in three months}. \\

\bottomrule
\end{tabular}
\caption{Examples of sentences with various types of grammatical errors.}
\label{tab:example2}
\end{table*}

\end{CJK*}

\subsection{Correct Sentences Collection}
To achieve perfect correct/ungrammatical sentence pairs for model training, we must make sure that the correct sentences are free of any grammatical errors as much as possible. However, collecting and annotating large-scale correct sentences are extremely time-consuming and expensive processes. To address this issue, we first accumulate massive amounts of high-quality Chinese sentences from public datasets~\cite{bright_xu_2019_3402023} such as the Chinese People’s Daily corpus, Chinese machine translation dataset, and Chinese wiki corpus as the raw corpora. Then we randomly selected 1,000 sentences from the raw corpora for human judgment by the annotators. The result that more than 97\% of the sentences are grammatically correct shows that, in a statistical sense, about 97\% of the sentences in the raw corpus do not contain grammatical errors. Further, to ensure the quality of our finally selected sentences, we only select sentences with perplexity values in the top 90\% (reverse order) as our original correct corpus. Finally, we acquire about 590K correct Chinese sentences from public corpora and guarantee their quality to a certain extent.

\subsection{Ungrammatical Sentence Generation}\label{sec:ungrammatical}
\begin{CJK*}{UTF8}{gbsn}
\begin{figure*}[h]
\centering
\subfigure[Improper Word Order] 
{ \label{fig:generation1} 
\includegraphics[width=1.0\columnwidth]{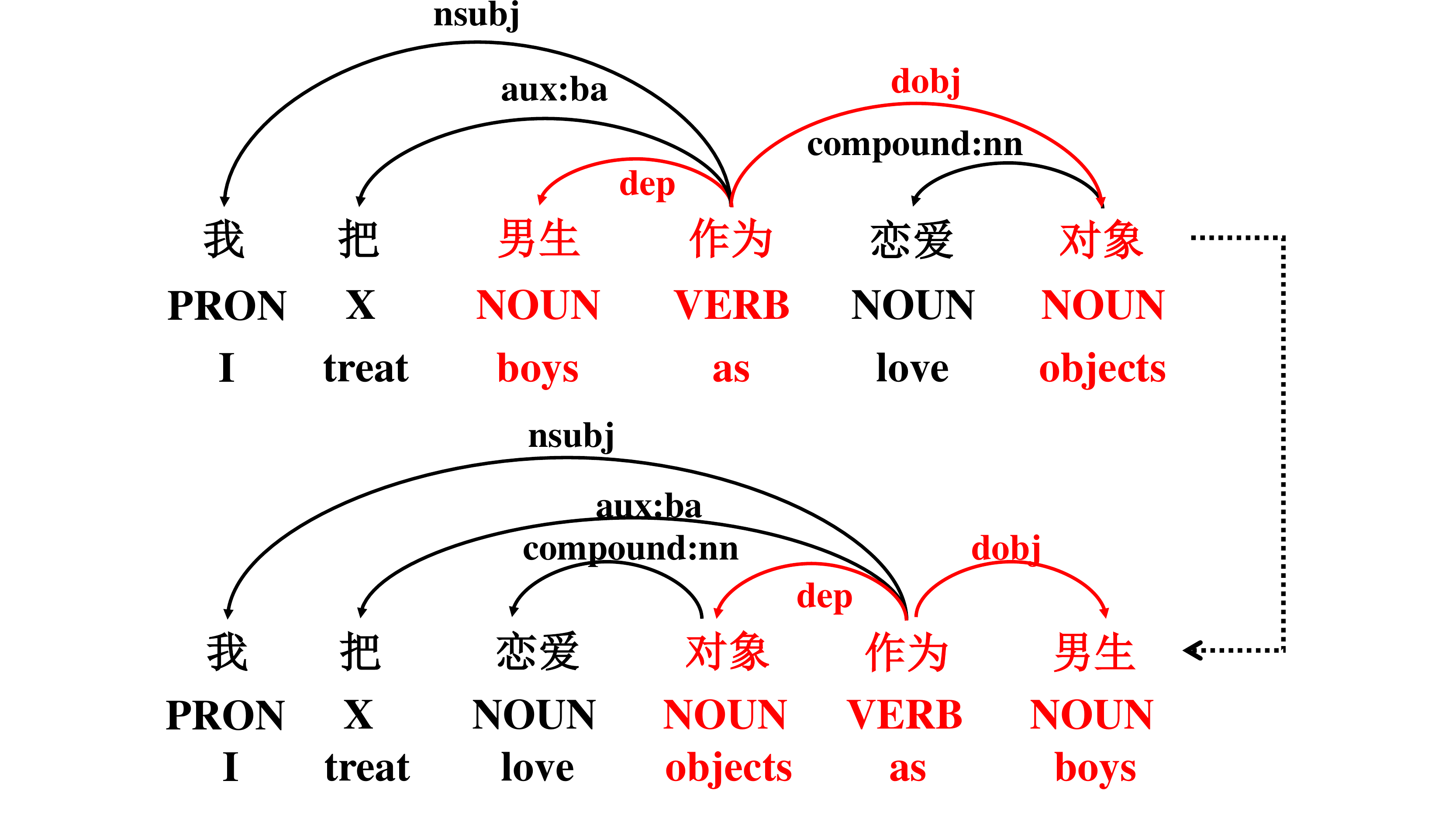} 
} 
\subfigure[Improper Logicality] 
{ \label{fig:generation6} 
\includegraphics[width=1.0\columnwidth]{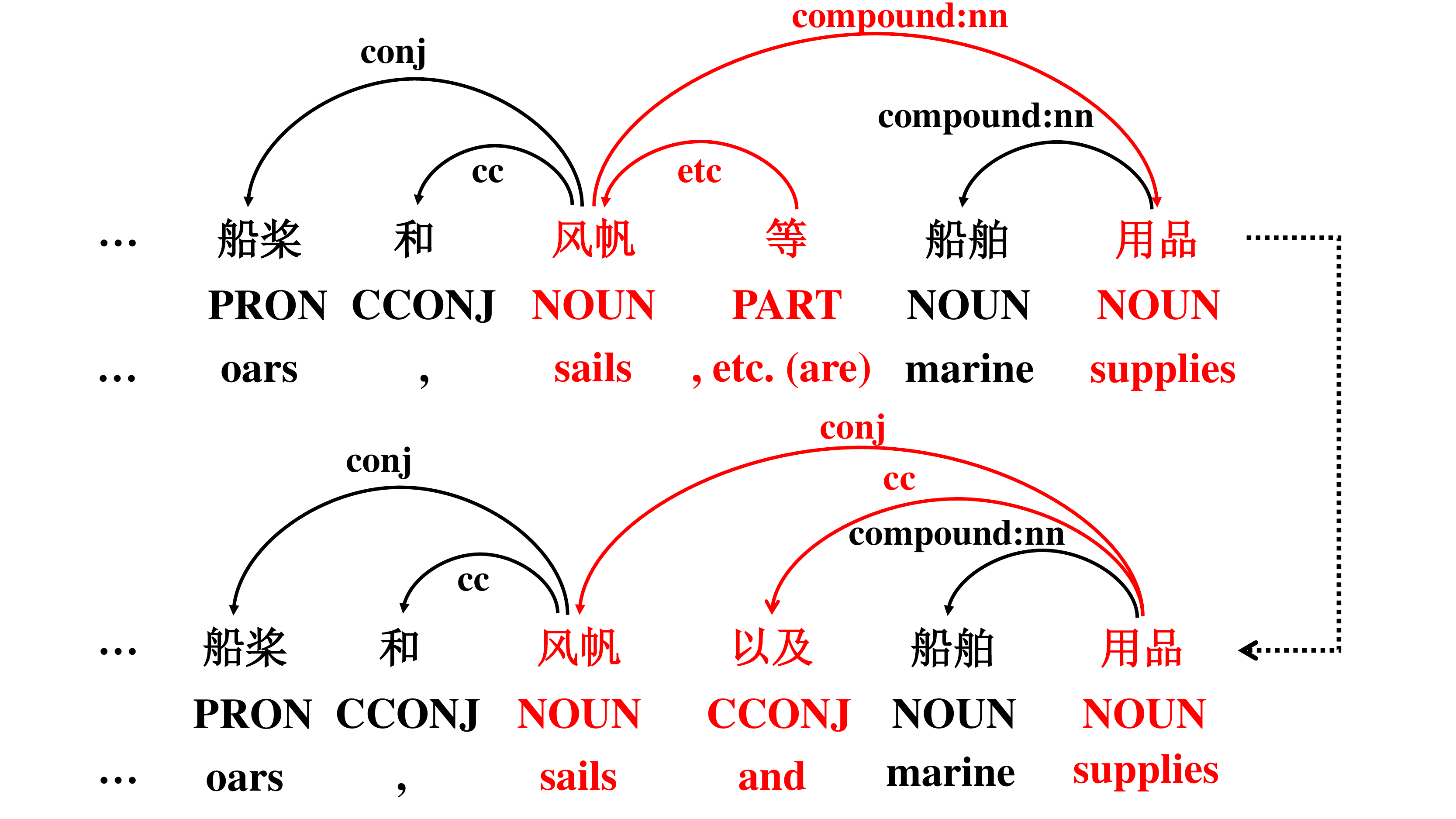} 
} 

\caption{Examples of generating ungrammatical sentences containing different types of errors. In the figures, NOUN represents “名词(noun)”, VERB represents “动词(verb)”, PRON represents “代词(pronoun)”, CCONJ represents “连词(conjunction)”, X represents “助动词(auxiliary verb)”.} 
\label{fig:generation} 
\end{figure*}
\end{CJK*}

After collecting correct sentences, we use the THULAC toolkit~\cite{sun2016thulac} for word segmentation and POS tagging, so that the six types of sentence components (subject, predicate, object, attribute, adverbial, complement) can be identified. Then for each category of grammatical errors, we design methods for constructing incorrect sentences from correct sentences according to grammatical rules and language conventions as follows:
\begin{enumerate}[(1)]
\item For \textbf{Structural Confusion}, we collect common correct sentence structures and mix two structures in a sentence to obtain an erroneous sentence that is also smooth and confusing.

\item For \textbf{Improper Logicality}, we collect some patterns with logical errors and match a pattern to a correct sentence to modify it to a wrong one.

\item For \textbf{Missing Component}, we pick a component from sentences to remove, or add words to the sentence to cover up a sentence component.

\item For \textbf{Redundant Component}, we select a notional word or meaningful connective in a sentence, and then insert a synonym before or after it to obtain an extra component.

\item For \textbf{Improper Collocation}, we summarize abundant common collocations between sentence components, and we match a collocation that appears in the sentence and randomly replace it with a wrong one.

\item For \textbf{Improper Word Order}, we switch the order of some important sentence components to make the sentence ungrammatical (e.g., reversing the order of related words or the order of an attribute and a head word).
\end{enumerate}

Due to the page limitation, in Figure~\ref{fig:generation}, we only present examples of generating sentences containing two types of errors, “Improper Word Order” and “Improper Logicality”, and the other examples are shown in Appendix~\ref{sec:otherexamples}.

Based on the above methods, we further design fine-grained grammatical error rules which are presented in Appendix~\ref{sec:finegrained_rules} and develop the corresponding algorithms\footnote{The implementation of the specific algorithm can refer to the supplementary materials we submitted.} to change correct sentences into ungrammatical sentences. 
Through this paradigm of automatically generating erroneous sentences of corresponding grammatical error types based on correct sentences, we obtain not only sentence pairs that can be used for model training, but also the labels of the grammatical error types of the corresponding sentence pairs.
Moreover, note that the rules mentioned in the above methods can be applied in combination to construct more complex erroneous instances which have multi grammatical errors in one sentence.

\subsection{Benchmark Collection and Annotation}\label{sec:benchmark}
Another core contribution of our work is that we construct the test dataset \DatasetName{} in which the grammatical errors are all made by native Chinese speakers in real-world scenarios. In this part, we describe how we collect and annotate \DatasetName{}.

Our benchmark focuses the native Chinese speakers' texts. To ensure that the language characteristics of \DatasetName{} are consistent with native Chinese speakers, we collect data from the following three real scenarios:
\begin{enumerate}[(1)]
    \item \textbf{Entrance examinations for secondary schools and universities}: In the various entrance examinations for Chinese students, there is a special type of question, that is, grammatical error diagnosis questions. These questions provide us with natural and real corpus with grammatical errors. Therefore, we collect the grammatical error questions of the middle school and university entrance examinations and the questions in the supporting teaching materials for the past 10 years for the construction of our test corpus. 
    \item \textbf{Recruitment examinations for government departments}: Similar to the entrance examinations for students, in China, the exams for civil servants employed by government agencies also include similar grammar questions. Different from the entrance exams, the question text of the civil service recruitment exams is closer to the domain of government official documents, while the questions of the entrance exam are closer to the domain of education. Likewise, we also collect texts of real test questions on civil service examinations and their auxiliary materials over the past 10 years.
    \item \textbf{Various Chinese news sites}: To guarantee the data size and domain diversity of our benchmark, we hired more than 20 data collectors and asked them to scour various mainstream Chinese news sites for sentences with grammatical errors. To ensure that they can find ungrammatical sentences more accurately, before they officially start their work, we have systematically taught them grammar knowledge, and asked them to pass the test of grammar error diagnosis questions set by us. 
\end{enumerate}

It is worth mentioning that the data collection process lasted for more than 6 months.
After obtaining ungrammatical corpus, we employ highly educated university students majoring in Chinese linguistics, who are familiar with Chinese grammatical rules, to check the collected corpus and annotate the corresponding error types according to our defined schema. The details of the annotation process are shown in Appendix~\ref{sec:annotation_process}.
We finally obtain 6,767 ungrammatical sentences and corresponding correction results to compose the \DatasetName{} benchmark.

\section{Data Analyses}
\subsection{Corpus Statistics}

Table~\ref{tab:statistics} reports the data statistics such as the number of sentences, average length, and average edit distance~\cite{levenshtein1966binary} of the corpus we constructed, including the automatically generated training data and the manually acquired test data.

\begin{table}[h]
\small
\centering
\begin{tabular}{l|cc}
\toprule
 & \MethodName{}-Train & \DatasetName{}-Test \\ \midrule
Number of Sentences & 591,404 & 6,767 \\
Erroneous Sentences & 591,404 & 6,496 \\
Number of References & 591,404 & 7,793 \\
Average Length (Char.) & 40.31 & 56.54 \\
Edit Distance (Char.) & 2.18 & 4.19 \\
References / Sentence & 1.00 & 1.20 \\
\bottomrule
\end{tabular}%
\caption{We report the statistics of the training corpus generated by \MethodName{} and the test data of \DatasetName{}.}
\label{tab:statistics}
\end{table}

\begin{table}[h]
\centering
\small
\begin{tabular}{l|p{0.65cm}p{0.65cm}p{0.65cm}p{0.65cm}}
\toprule
 & Replace & Insert & Delete & Total \\
\midrule
Structural Confusion & 0.44 & 0.87 & 1.55 & 2.86 \\
Improper Logicality & 0.60 & 0.90 & 2.06 & 3.56 \\
Missing Component & 0.10 & 2.10 & 0.42 & 2.62 \\
Redundant Component & 0.09 & 0.06 & 2.00 & 2.15 \\
Improper Collocation & 1.28 & 0.79 & 0.79 & 2.86 \\
Improper Word Order & 0.46 & 5.37 & 5.42 & 11.24 \\
\bottomrule
\end{tabular}
\caption{Average edit distance for sentences containing various types of grammatical errors. Edit operations include Replace, Insert and Delete.}
\label{tab:distance}
\end{table}

In addition, we perform further analysis on the test samples in \DatasetName{}.
Specifically, we analyze the character-level edit distance required to correct sentences for different types of errors. Note that a sentence with multiple errors will be counted in the corresponding error types repeatedly.
From Table~\ref{tab:distance}, we see that the average edit distance for sentences with most types is between 2 and 4, while that for sentences with “Improper Word Order” is as high as 11.24. It indicates that such error requires significant adjustments to sentence structure, which is a challenge for CGEC models.

\subsection{Human Evaluation}

To verify that our \DatasetName{} more closely matches the language style of Chinese native speakers and the grammatical errors that native speakers would make, we conduct a human evaluation experiment.

Specifically, we randomly sample 300 correct sentences and 300 incorrect sentences from the test sets of NLPCC, CGED, and \DatasetName{} respectively.
We carefully check these sentences to ensure that the correct and incorrect labels of them are correct from the perspective of native Chinese speakers.

Then we invite 3 Chinese native speakers to do the following two annotation task for these sentences: 
(1) Annotators should determine whether a sentence contains grammatical errors or not. 
(2) Annotators should determine whether the language style of a sentence matches that of a native speaker, giving a score of 0, 1, or 2 depending on the degree of match. 
We compare the results of whether the sentences contain grammatical errors judged by the annotators with the corresponding ground truth, and calculate Precision, Recall, and F$_1$ of binary classification made by annotators. The results of 3 annotators are averaged and reported in Table~\ref{tab:human}.

\begin{table}[ht]
\small
\centering
\begin{tabular}{l|cccc}
\toprule
 & Pre & Rec & F$_{1}$ & Score \\ \midrule
NLPCC & 78.57 & 82.76 & 80.61 & 0.92 \\
CGED & 95.00 & 90.48 & 92.68 & 0.85 \\
\DatasetName{} & 72.86 & 68.00 & 70.34 & 1.78 \\ \bottomrule
\end{tabular}%
\caption{Results of human evaluation experiment. Precision, Recall, and F$_1$ are metrics of annotators distinguishing whether sentences contain grammatical errors, and the Score represents the extent to which annotators judge that sentences conform to the language habits of native speakers (0, 1, 2).}
\label{tab:human}
\end{table}

The results demonstrate that:
(1) Compared to NLPCC and CGED datasets, the test dataset of \DatasetName{} obtains a significantly higher language style score, suggesting that the sentences in \DatasetName{} are more in line with the language habits of native speakers.
(2) Annotators who are Chinese native speakers have more difficulty distinguishing whether the sentences from our benchmark contain grammatical errors, indicating that the grammatical errors in our benchmark are more likely to be made by native speakers in their everyday writing.

\section{Experiments}
\subsection{Experimental Setup}
\begin{table*}[t]
\small
\centering
\begin{tabular}{l|l|ccc}
\toprule
\multicolumn{1}{c}{} & \multicolumn{1}{c}{} & Pre & Rec & F$_{0.5}$ \\ \midrule
\multirow{4}{*}{Transformer~\cite{DBLP:conf/nips/VaswaniSPUJGKP17}} & Data Aug.(1000K) & 3.50 & 1.49 & 2.76 \\
 & Lang8(1220K) & 8.22 & 1.04 & 3.44 \\
 & Lang8+HSK(1377K) & 5.91 & 0.79 & 2.57 \\
 & \MethodName{}(591K) & 17.19 & 6.20 & 12.69 \\ 
 & Lang8+\MethodName{}(1811K) & 26.75 & 5.89 & 15.66 \\
\midrule
\multirow{4}{*}{GECToR-Chinese~\cite{DBLP:journals/corr/abs-2204-10994}} & Data Aug.(1000K) & 4.35 & 1.85 & 3.42 \\
 & Lang8(1220K) & 20.77 & 6.97 & 14.88 \\
 & Lang8+HSK(1377K) & 22.01 & 8.73 & 16.88 \\
 & \MethodName{}(591K) & 23.25 & 11.03 & 19.04 \\ 
 & Lang8+\MethodName{}(1811K) & 27.71 & 12.19 & 22.09 \\
\bottomrule
\end{tabular}%
\caption{Experimental results on \DatasetName{} benchmark. The number in parentheses represents the number of sentences in the corresponding dataset.}
\label{tab:benchmark}
\end{table*}

We evaluate the performance of two mainstream CGEC methods on our benchmark, including a seq2seq model and a seq2edit model:

\paragraph{Transformer}~\cite{DBLP:conf/nips/VaswaniSPUJGKP17} is a widely used model with the encoder-decoder structure and it is usually utilized to resolve seq2seq tasks. Following previous work~\cite{DBLP:conf/nlpcc/FuHD18, wang2020chinese, tang2021chinese} which treats CGEC as a monolingual translation task and employs the Transformer on the NLPCC, we implement and train a Transformer as our seq2seq model.
\paragraph{GECToR-Chinese}~\cite{DBLP:journals/corr/abs-2204-10994} is a seq2edit model, which treats CGEC as a sequence labeling task and predicts the modification label, including insertion, deletion, and substitution~\cite{malmi-etal-2019-encode}, for each token of the sentence. The model adopts GECToR~\cite{omelianchuk-etal-2020-gector} and enhances it by applying 
a pretrained language model as the encoder.

Following the NLPCC shared task~\cite{DBLP:conf/nlpcc/ZhaoJS018}, we employ word-level \textbf{MaxMatch (M$_2$)} Scorer~\cite{dahlmeier-ng-2012-better} for evaluation, which computes Precision, Recall, and F$_{0.5}$ between the gold edit set and the system edit set.

We train the model using the Lang8~\cite{DBLP:conf/nlpcc/ZhaoJS018} dataset which is the official training dataset of NLPCC, HSK~\cite{cui2011principles, zhang2013design} dataset, and the training dataset generated by \MethodName{}. In addition, we construct a training dataset by data augmentation~\cite{wang2020chinese, tang2021chinese} for the comparison between our proposed \MethodName{} and traditional data augmentation methods. After the training stage, we test the trained models on \DatasetName{} benchmark.
Additionally, the implementation details of our experiments are presented in Appendix~\ref{sec:implementation}.

\subsection{Experimental Results}
The experimental results shown in Table~\ref{tab:benchmark} demonstrate that:
(1) Existing models do not perform well on our benchmark, with F$_{0.5}$ being below 25, revealing that \DatasetName{} benchmark is challenging.
(2) For both Transformer and GECToR-Chinese, models trained on \MethodName{} training dataset outperform models trained on other datasets, among which Transformer trained on \MethodName{} performs far better than that trained on other datasets. It reflects that the training data generated by our proposed \MethodName{} effectively assists models to correct grammatical errors made by Chinese native speakers.
(3) \MethodName{} and Lang8 dataset can be jointly employed to train the model and further improve the performance of the model, which demonstrates that the dataset constructed by \MethodName{} is compatible with existing datasets without conflicts.

\subsection{Analysis of Generalization Ability}
\begin{table}[t]
\centering
\small
\begin{tabular}{l|p{0.4cm}p{0.4cm}p{0.4cm}p{0.4cm}}
\toprule
 \multicolumn{1}{c}{} & Pre & Rec & F$_{0.5}$ & \multicolumn{1}{l}{$\Delta$F$_{0.5}$} \\ \midrule
 No Pretrained & 34.17 & 13.41 & 26.09 & - \\
 Data Aug. (1000K) & 41.49 & 14.48 & 30.21 & +4.12 \\
 Data Aug. (1600K) & 42.30 & 15.94 & 31.79 & +5.70 \\
 \MethodName{} (591K)  & 38.24 & 16.64 & 30.36 & +4.27 \\
 \MethodName{} + Data Aug. (1591K) & 41.73 & 17.02 & 32.34 & +6.25 \\ 
 \bottomrule
\end{tabular}%
\caption{Experimental results of Transformer on NLPCC benchmark.}
\label{tab:nlpcc2018}
\end{table}

Furthermore, to verify the generalizability of the data generated by \MethodName{}, we conduct an experiment on the NLPCC dataset.
To be specific, before training with the Lang8 dataset, we pretrain the model using \MethodName{} training corpus. The compared baseline models include: the Transformer without pretraining and the Transformer pretrained on a corpus constructed by traditional data augmentation approach.
Following previous work~\cite{wang2020chinese, tang2021chinese}, we implement the following data augmentation approach. After a correct sentence is segmented into words, the following operations are performed for each word in the sentence according to different probabilities: 70\% of no modification, 10\% of inserting a random word before this word, 10\% of replacing the word with a random word, and 10\% of deleting this word.

The results in Table~\ref{tab:nlpcc2018} show that:
(1) Pretraining the model with \MethodName{} improves the model's performance on the NLPCC test dataset, and the improvement is greater than pretraining the model with a larger corpus constructed by traditional data augmentation, which reflects the superiority of our \MethodName{} over traditional data augmentation methods on the CGEC task. 
(2) Compared with directly increasing the size of corpus constructed by data augmentation, our \MethodName{} and data augmentation approach can be combined to achieve better results.
To summarize, the training data constructed by our proposed \MethodName{} brings significant and stable improvements to the model on both \DatasetName{}, which focuses on native speaker errors, and NLPCC, which focuses on foreign learner errors. This phenomenon proves the good generalization ability of the data constructed by \MethodName{}.

\subsection{Case Study and Fine-grained Analysis}
\begin{CJK*}{UTF8}{gbsn}
\begin{table}[ht]
\small
\centering
\begin{tabular}{c|l}
\toprule
\multirow{2}{*}{Incorrect} & 其对象主要是\textcolor{red}{面向}低收入家庭 \\ 
 & Its target is mainly \textcolor{red}{for} low-income families \\
\midrule
\multirow{2}{*}{Correct} & 其对象主要是低收入家庭 \\
 & Its target is mainly low-income families \\
\midrule
\textcolor{red}{\makecell[c]{Model\\(Lang8)}} & 其对象主要是\textcolor{red}{面向}低收入家庭 \\
\midrule
\textcolor{blue}{\makecell[c]{Model\\(\MethodName{})}} & 其对象主要是低收入家庭 \\
\midrule
Type & Structural Confusion \\
\bottomrule

\toprule
\multirow{2}{*}{Incorrect} & 请\textcolor{red}{把这件事你不要}放在心上 \\ 
 & I hope \textcolor{red}{this matter you do not} keep in mind \\
\midrule
\multirow{2}{*}{Correct} & 请你不要把这件事放在心上 \\
 & I hope you do not keep this matter in mind \\
\midrule
\multirow{2}{*}{\textcolor{red}{\makecell[c]{Model\\(Lang8)}}} & 请把这件事放在心上 \\
 & I hope you keep this matter in mind \\
\midrule
\textcolor{orange}{\makecell[c]{Model\\(\MethodName{})}} & 请不要把这件事\textcolor{red}{你}放在心上 \\
\midrule
Type & Improper Word Order \\
\bottomrule
\end{tabular}
\caption{Cases of models trained on different datasets correcting ungrammatical sentences. The \textcolor{red}{red}/\textcolor{blue}{blue}/\textcolor{orange}{orange} color in the first column represents that the prediction result of the corresponding model is \textcolor{red}{wrong}/\textcolor{blue}{correct}/\textcolor{orange}{approximately correct}.}
\label{tab:case}
\end{table}
\end{CJK*}

\begin{CJK*}{UTF8}{gbsn}
Table~\ref{tab:case} illustrates two cases of models trained on different datasets correcting erroneous sentences in \DatasetName{}. 
The first sentence contains an error of “Structural Confusion”, where “对象...面向(target...for)” is a mixture of two different sentence patterns. The model trained on the Lang8 dataset cannot recognize such an error, while the model trained on \MethodName{} perceive the error and correct the sentence properly. 
The second sentence involves an error of “Improper Word Order”, reversing the order of “把这件事(this matter)” and “你不要(you do not)”. The error is so difficult for models that the model trained on Lang8 removes “你不要”, which makes the corrected sentence grammatical but entirely changes the original sentence meaning. Meanwhile, the model trained on \MethodName{} does not correct it completely, but identifies the sentence including an error of “Improper Word Order” and makes a partially appropriate correction.
From these two cases, it can be further revealed that:
(1) The distribution of grammatical errors in the Lang8 dataset differs significantly from the distribution of grammatical errors that occur in the everyday writing of native speakers.
(2) The model trained with \MethodName{} can better identify and correct the grammatical errors of native speakers.
\end{CJK*}

\begin{table}[t]
\centering
\small
\begin{tabular}{l|ccc}
\toprule
 & Pre & Rec & F$_{0.5}$ \\
\midrule
Structural Confusion & 37.14 & 23.53 & 33.28 \\
Improper Logicality & 31.63 & 19.17 & 27.99 \\
Missing Component & 9.48 & 4.00 & 7.44 \\
Redundant Component & 27.75 & 15.90 & 24.15 \\
Improper Collocation & 7.82 & 3.30 & 6.13 \\
Improper Word Order & 19.82 & 5.65 & 13.20 \\
\bottomrule
\end{tabular}
\caption{Model performance on different types of grammatical errors.}
\label{tab:types}
\end{table}

Additionally, we also analyze the performance of the model on sentences with fine-grained grammatical error types. The results of Table~\ref{tab:types} illustrate that the model performs particularly poorly in dealing with three types of error, i.e., “Missing Component”, “Improper Collocation” and “Improper Word Order”, which reveals that existing methods have insufficient understanding of common grammatical collocations and insufficient knowledge of macroscopic sentence structure.
\emph{We suggest that future research for the CGEC task should be more refined and pay more attention to the connections among each individual component of sentences and the overall structure of sentences.}

\section{Conclusion}
In this paper, we propose the \MethodName{} based on linguistic rules to automatically generate ungrammatical sentences from correct texts for obtaining large-scale CGEC training corpus. Besides, we collect sentences with grammatical errors written by Chinese native speakers and construct a more challenging CGEC benchmark \DatasetName{}. Experimental results and detailed analyses indicate that \DatasetName{} is consistent with the language habits of Chinese native speakers, and the training data generated by \MethodName{} effectively improves the model performance on \DatasetName{} and other mainstream benchmarks. We hope our work provides better resources and a new direction for future CGEC research.

\section{Limitations}
\subsection{Limitation of Language}
This work focuses on the Chinese Grammatical Error Correction (CGEC) task. The grammatical error classification and ungrammatical sentence generation approach \MethodName{} involved in this paper are based on Chinese linguistic rules and cannot be directly extended to other languages such as English. However, we believe that a linguistic perspective can also be introduced in GEC tasks of other languages for more in-depth analysis and exploration.

\subsection{Limitation of Experiments}
We conduct experiments on CGEC employing two mainstream approaches, including Transformer and GECToR-Chinese models. Limited by hardware resources, the total number of parameters of the two models we construct are around 54M and 108M respectively, and we do not try to build larger models or utilize model ensemble. Future work can construct larger models and use more tricks to improve the performance on our benchmark.

\section{Ethical Considerations}
In this paper, we propose \MethodName{} that automatically generate large-scale and high-quality CGEC training data based on our designed linguistic rules. Additionally, we present a human-annotated benchmark, \DatasetName{}, which focuses on the grammatical errors made by native Chinese speakers. We describe the details of our proposed corpus generation method and our constructed benchmark in the main text. It is worth noting that in the process of generating training data and labeling test data, all the original corpora used are from publicly available datasets or resources on the legitimate websites, and no sensitive data is involved. Additionally, for CGEC itself, it is a task that has been concerned for a long time and has a wide range of application scenarios. Our work focuses on an important issue that has been overlooked in previous CGEC work, namely grammatical errors made by native Chinese speakers. Therefore, \DatasetName{} is likely to directly facilitate the research of CGEC, and further enhance the ability of future CGEC models to deal with hard errors made by native speakers. 

\section*{Acknowledgement}
This research is supported by National Natural Science Foundation of China (Grant No.62276154 and 62011540405), Beijing Academy of Artificial Intelligence (BAAI), the Natural Science Foundation of Guangdong Province (Grant No. 2021A1515012640), Basic Research Fund of Shenzhen City (Grant No. JCYJ20210324120012033 and JCYJ20190813165003837), and Overseas Cooperation Research Fund of Tsinghua Shenzhen International Graduate School  (Grant No. HW2021008).

\bibliography{anthology,custom}
\bibliographystyle{acl_natbib}

\clearpage
\appendix

\section{Appendix}
\subsection{Other Examples of Ungrammatical Sentence Generation}
\label{sec:otherexamples}
\begin{CJK*}{UTF8}{gbsn}

\begin{figure*}[h]
\centering
\subfigure[Structural Confusion] 
{ \label{fig:generation2} 
\includegraphics[width=1.0\columnwidth]{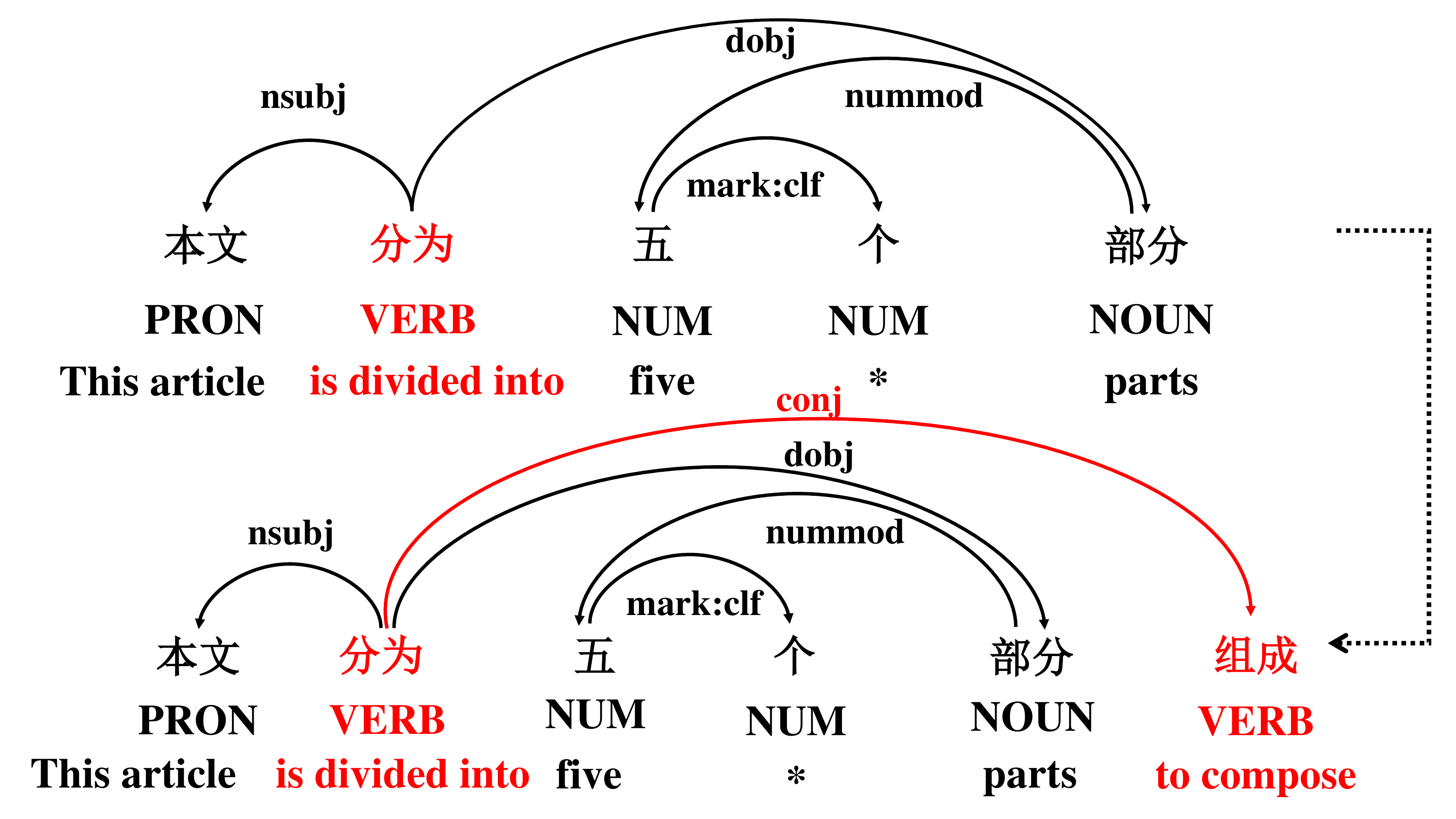} 
} 
\subfigure[Redundant Component] 
{ \label{fig:generation3} 
\includegraphics[width=1.0\columnwidth]{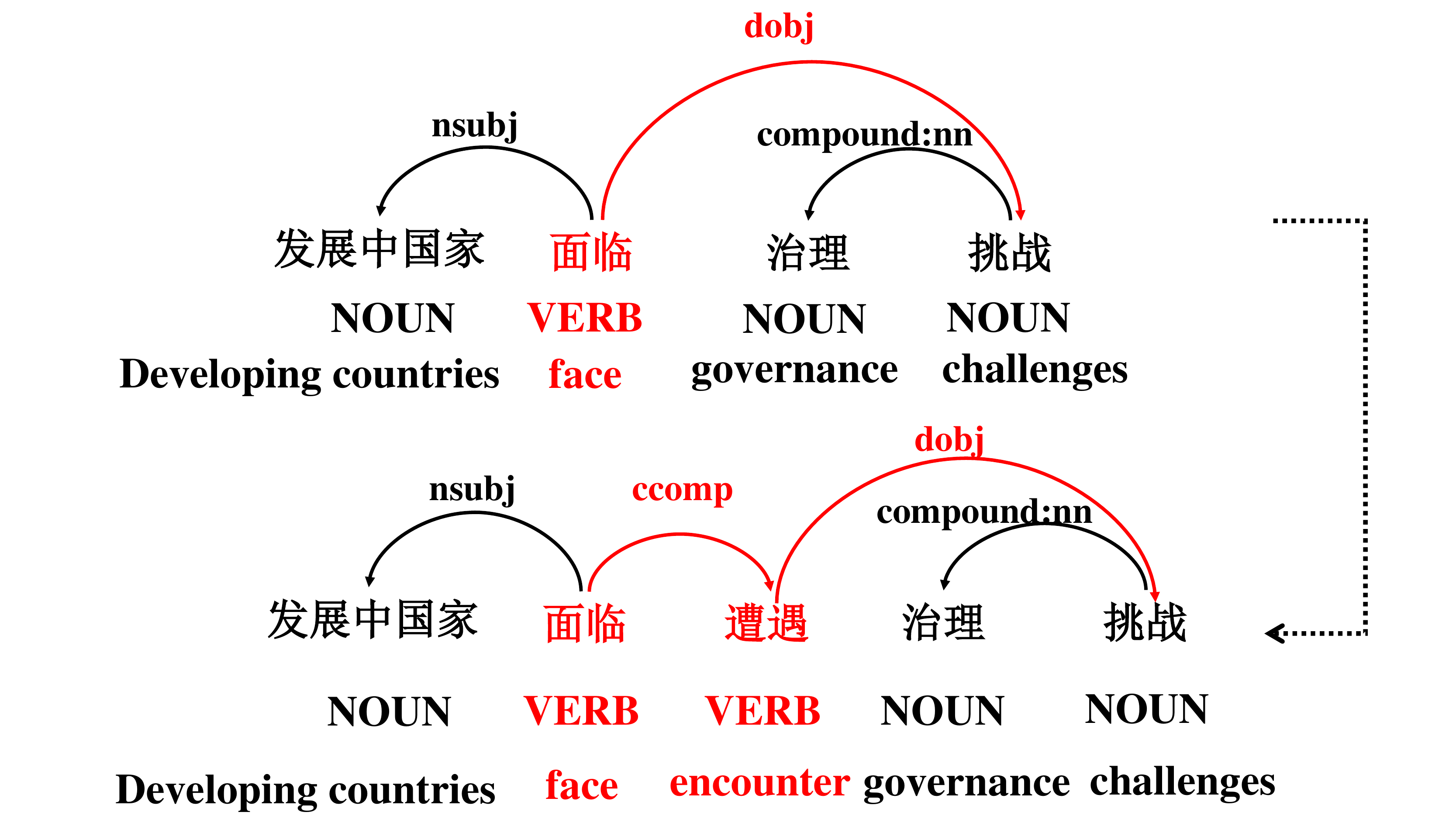} 
}
\subfigure[Missing Component] 
{ \label{fig:generation4} 
\includegraphics[width=1.0\columnwidth]{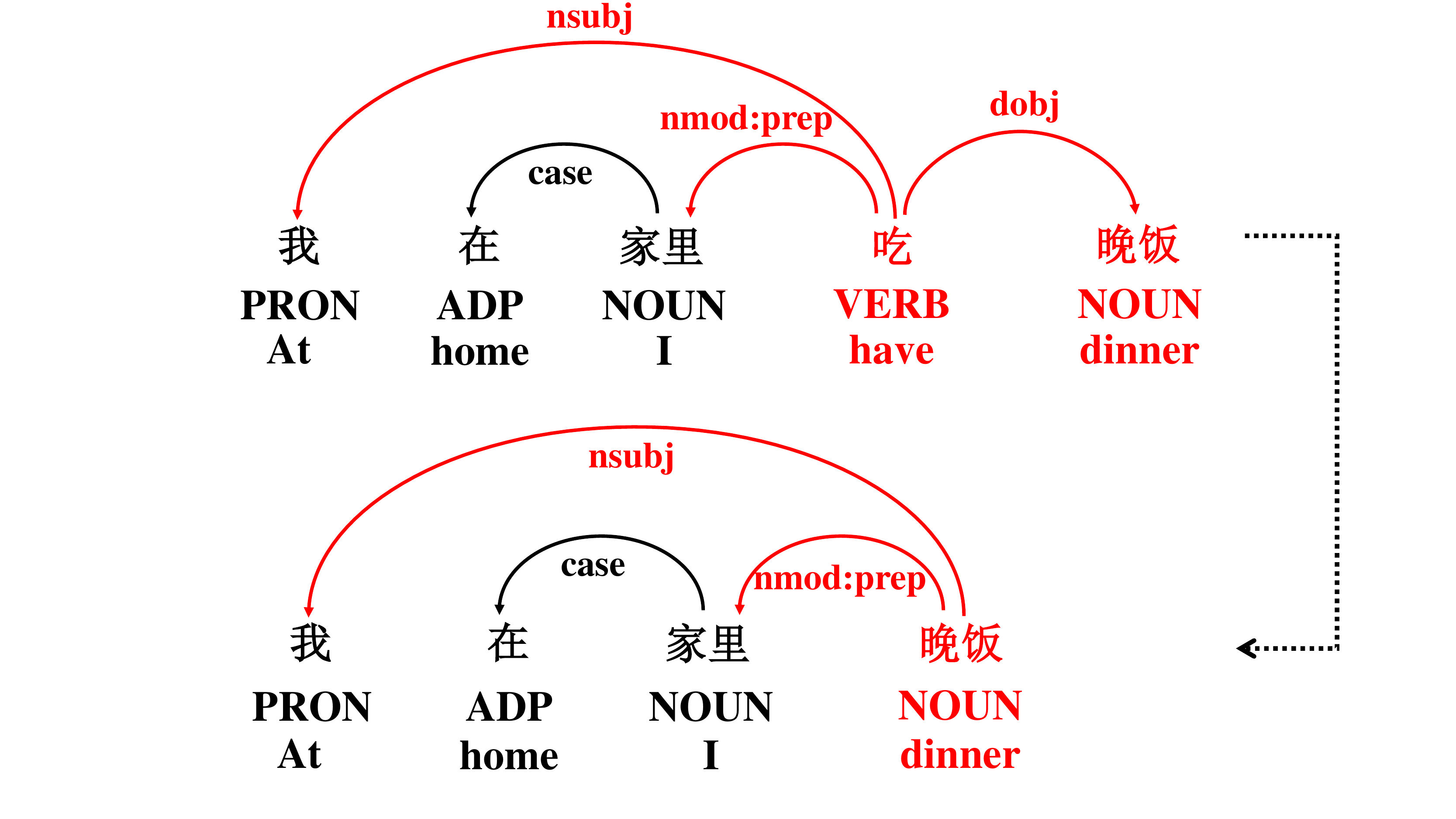} 
}
\subfigure[Improper Collocation] 
{ \label{fig:generation5} 
\includegraphics[width=1.0\columnwidth]{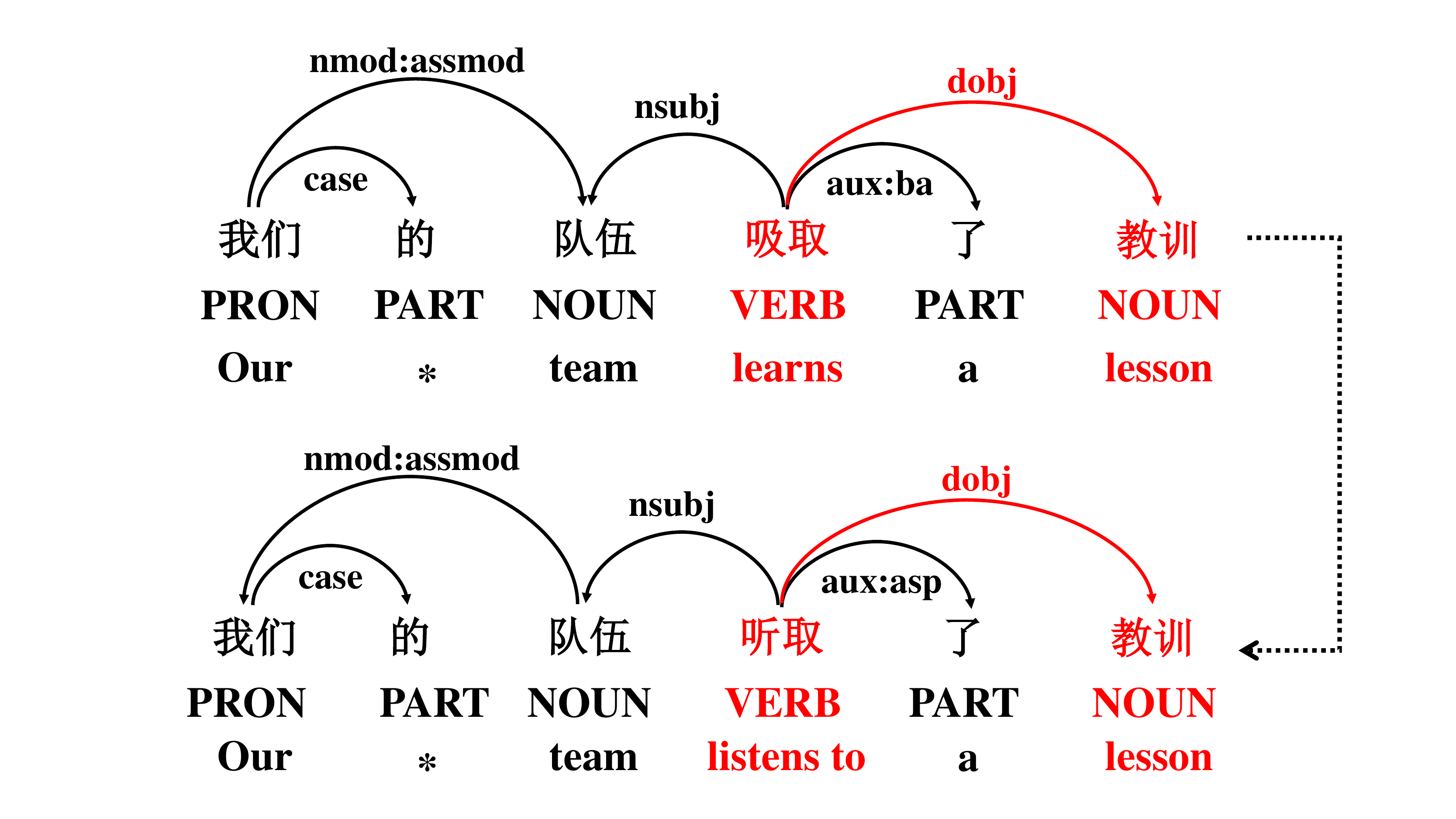} 
}
\caption{Other Examples of generating ungrammatical sentences containing different types of errors.} 
\label{fig:generation-others} 
\end{figure*}

Figure~\ref{fig:generation-others} presents some examples of generating ungrammatical sentences that are not mentioned in the body of this paper. In the figures, NOUN represents “名词(noun)”, VERB represents “动词(verb)”, PRON represents “代词(pronoun)”, CCONJ represents “连词(conjunction)”, ADP represents “介词(preposition)”, X represents “助动词(auxiliary verb)”, PART represents “助词(particle/auxiliary word)”, NUM represents “数量词(numeral)”.
\end{CJK*}

\subsection{Fine-grained Grammatical Error Rules}
\label{sec:finegrained_rules}

\begin{CJK*}{UTF8}{gbsn}
\begin{figure*}[ht]
\centering
\includegraphics[width=2.0\columnwidth]{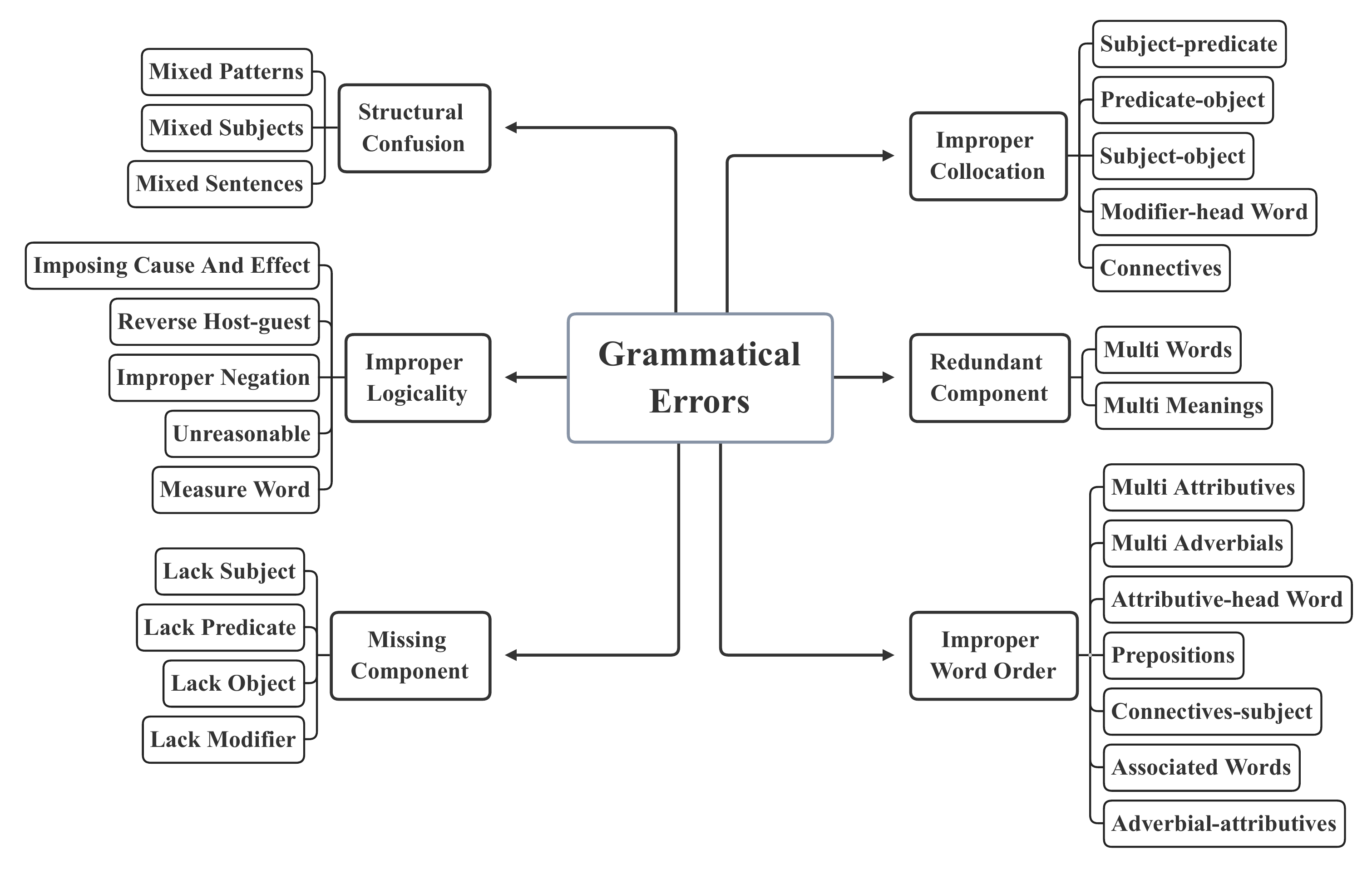} 
\caption{An overview of the fine-grained classification of different types of grammatical errors.} 
\label{fig:errors} 
\end{figure*}

As described in Section~\ref{sec:ungrammatical}, we design rules for generating sentences containing corresponding errors for each of the 6 types of grammatical errors. Moreover, each of these 6 types of errors can be divided into several fine-grained error types. As shown in Figure~\ref{fig:errors}, these grammatical errors can be separated into a total of 26 fine-grained types of error. In fact, for each fine-grained type of errors, we consider their characteristics and design the corresponding rules. These fine-grained grammatical error rules are explained as follows:

\begin{enumerate}[(1)]
    \item For \textbf{Structural Confusion}, there are 3 fine-grained error types.
    \begin{itemize}
        \item \textbf{Mixed Patterns (句式杂糅)} means confusion in sentence structure caused by mixing different expressions of the same meaning together.
        \item \textbf{Mixed Subjects (中途易辙)} means a sentence has two subjects but without any connectives.
        \item \textbf{Mixed Sentences (两句合一)} refers to the inappropriate merging of two sentences, that is, the structure of one sentence is complete, but the last part of the sentence is used as the beginning of the other sentence.
    \end{itemize}
    \item For \textbf{Improper Logicality}, there are 5 fine-grained error types.
    \begin{itemize}
        \item \textbf{Measure Word (数词不当)} refers to the exact and approximate measure words are used in a sentence simultaneously.
        \item \textbf{Unreasonable (不合事理)} means the sentence doesn't conform to formal or moral logic such as conceptual judgments.
        \item \textbf{Improper Negation (否定失当)} means that the sentence uses multiple negation, making the sentence meaning reversed.
        \item \textbf{Reverse Host-guest (主客倒置)} means that the position of the host and guest described in a sentence are reversed.
        \item \textbf{Imposing Cause and Effect (强加因果)} means that the sentence forces a connection between two events that are not directly causally linked.
    \end{itemize}
    \item For \textbf{Missing Component}, there are 4 fine-grained error types.
    \begin{itemize}
        \item \textbf{Lack Subject (主语残缺)} means that the sentence does not have a subject.
        \item \textbf{Lack Predicate (谓语残缺)} means that the sentence does not have a predicate.
        \item \textbf{Lack Object (宾语残缺)} means that the sentence does not have an object.
        \item \textbf{Lack Modifier (修饰成分残缺)} refers to sentences that lack necessary modifiers.
    \end{itemize}
    \item For \textbf{Redundant Component}, there are 2 fine-grained error types.
    \begin{itemize}
        \item \textbf{Multi Words (堆砌词语)} refers to the use of two or more words in a sentence where one word would have been sufficient, resulting in redundancy.
        \item \textbf{Multi Meanings (语义重复)} means that two or more words of the same meaning are used in a sentence or that the latter word already contains the meaning of the former word.
    \end{itemize}
    \item For \textbf{Improper Collocation}, there are 5 fine-grained error types.
    \begin{itemize}
        \item \textbf{Subject-predicate (主谓搭配不当)} means the inappropriate pairing of subject and predicate in a sentence.
        \item \textbf{Predicate-object (动宾搭配不当)} and \textbf{Subject-object (主宾搭配不当)} are almost the same as \textbf{Subject-predicate}.
        \item \textbf{Modifier-head Word (修饰语中心语搭配不当)} means the modifier does not correctly modify the head word in a sentence.
        \item \textbf{Connectives (关联词语搭配不当)} refers to the incorrect collocation of connectives used in a sentence.
    \end{itemize}
    \item For \textbf{Improper Word Order}, there are 7 fine-grained error types.
    \begin{itemize}
        \item \textbf{Multi Attributives (多重定语)} means that the multi-attributives' order is wrong.
        \item \textbf{Multi Adverbials (多重状语)} means that the order of multi-adverbials is incorrect.
        \item \textbf{Attributive-head Word (定语与中心语次序不当)} means that the positions of attributive and head word are reversed.
        \item \textbf{Prepositions (介词短语位置不当)} refers to wrong positions of prepositions.
        \item \textbf{Connectives-subject (关联词位置不当)} refers to the improper location of connectives in clauses.
        \item \textbf{Associated Words (副词位置不当)} means that the associated words or phrases are not in the right order.
        \item \textbf{Adverbial-attributives (定语状语次序不当)} means that attributives and adverbs misuse each other.
    \end{itemize}
\end{enumerate}

\subsection{Details of \DatasetName{} Annotation Process}
\label{sec:annotation_process}
The resources of \DatasetName{} mainly come from three aspects, namely entrance examinations, recruitment examinations, and various Chinese news sites. Note that the data from entrance/recruitment examination scenarios include specific error types, correct sentences and even analysis of the reasons for the errors, because these data are all from real exam questions. Therefore, our annotation work is mainly carried out on data from news websites. To be specific, our annotation team consists of 5 annotators and 1 senior annotation referee. Our annotation workflow mainly consists of two parts:
\begin{enumerate}[(1)]
    \item To get the effective grammatical error labels, we ask each annotator to give each sample the type of grammatical error he think the sample has. So for each sample, we will get 5 preliminary annotation results. We will select the one with the most occurrences among these five results as the final labeling result for this sample. If there are multiple error types with the same number of occurrences, the senior referee will decide the final error type for this sample based on his profound knowledge.
    \item To get the reasonable reference correct sentences, we require each annotator to rewrite every sentence and give as many rewrites as they feel feasible. Finally, we will select the rewriting results that appear twice or more as the reference standard sentences for the erroneous sentence. If the rewriting results given by the 5 annotators differ from each other for a sample, we will ignore this sample directly after the senior judge has reviewed it.
\end{enumerate}

Additionally, considering that the process of labeling grammatical error types is essentially a multi-label classification process, we use the Fleiss' kappa~\cite{fleiss1971measuring} to verify the annotator agreement of labeling grammatical error types. The final Fleiss' kappa score is 0.823, which indicates that our annotation results can be regarded as “almost perfect agreement”~\cite{landis1977measurement}.

\subsection{Implementation Details}
\label{sec:implementation}
For the Transformer, we implement the code using PyTorch~\cite{DBLP:conf/nips/PaszkeGMLBCKLGA19}. The word embedding matrix shares weight between the source side and the target side, and the embedding dimension is set to 512. Both the encoder and decoder of the Transformer contain 6 layers, and each layer contains 8 attention heads. The maximum length of the input is set to 200, and sentences longer than 200 are truncated. We train the model with the AdamW~\cite{DBLP:journals/corr/abs-1711-05101} optimizer for 20 epochs. The learning rate and dropout rate are set to 5e-4 and 0.1 respectively. In the inference phase, we use Beam Search as the decoding method, with the number of beams being 8.

For the GECToR-Chinese, we utilize the code provided by the original authors~\cite{DBLP:journals/corr/abs-2204-10994}.
Due to the limitation of hardware resources, we select MacBERT(Base)~\cite{cui-etal-2020-revisiting} as its encoder, unlike the original code using StructBERT~\cite{DBLP:conf/iclr/0225BYWXBPS20}. The maximum sentence length is also set to 200. Adam~\cite{DBLP:journals/corr/KingmaB14} is employed as the model optimizer. The training process is divided into two stages: in the first stage the parameters of the encoder are frozen and the model is trained for 2 epochs with a learning rate set to 1e-3, and in the second stage the full parameters are tuned and the model is trained for 20 epochs with a learning rate set to 2e-5.

\end{CJK*}
\end{document}